\RequirePackage{fix-cm}
\documentclass[twocolumn]{svjour3}          
\smartqed  
\usepackage{graphicx}
\journalname{Signal Processing: Image Communication}

\usepackage[export]{adjustbox}

\usepackage[hyphens]{url}
\usepackage[english]{babel}
\usepackage[utf8]{inputenc}

\usepackage[gen]{eurosym}
\begin{document}

\title{Information Extraction from Scanned Invoice Images using Text
Analysis and Layout Features}

\author{H.~T.~Ha\and A.~Hor\'{a}k}

\authorrunning{}

\institute{Natural Language Processing Centre\\
Faculty of Informatics, Masaryk University\\ 
Botanick\'{a} 68a 602 00 Brno, Czech Republic\\
\email{xha1@fi.muni.cz},
\email{hales@fi.muni.cz}}

\date{Received: date / Accepted: date}

\maketitle

\sloppy
\raggedbottom

\begin{abstract}

While storing invoice content as metadata to avoid paper document
processing may be the future trend, almost all of daily issued
invoices are still printed on paper or generated in digital formats
such as PDFs. In this paper, we introduce the OCRMiner system for
information extraction from scanned document images which is based on
text analysis techniques in combination with layout features to
extract indexing metadata of (semi-)structured documents. The system
is designed to process the document in a similar way a human reader
uses, i.e.\ to employ different layout and text attributes in
a coordinated decision. The system consists of a set of interconnected
modules that start with (possibly erroneous) character-based output
from a standard OCR system and allow to apply different techniques and
to expand the extracted knowledge at each step.  Using an open source
OCR, the system is able to recover the invoice data in 90\% for
English and in 88\% for the Czech set.

\end{abstract}

\section{Introduction}

Automatic invoice processing systems gain significant interest of
large companies who deal with enormous numbers of invoices each day,
due to not only their legal value requiring them to be stored for
years but also for economic reasons. Cristani et
al in~\cite{cristani2018future} offer a comparison of \EUR{9} per manually
processed invoice and \EUR{2} per automated processing of one invoice
based on surveys in 2004 and 2003 respectively. A 2016 report by the
Institute of Finance and Management~\cite{iofm2016} suggested that the
average cost to process an invoice was \$12.90.

Several challenges on recognition and extraction of key texts from
scanned receipts and invoices have been organized recently, e.g.\ the
Robust Reading Challenge on Scanned Receipt OCR and Information
Extraction (SROIE) at ICDAR 2019~\cite{jaume2019funsd} or the Mobile-Captured
Image Document Recognition for Vietnamese Receipts at
RIVF2021~\cite{vu2021mc}.
Still, annotated benchmark invoice data\-sets are not generally
available due to confidential information, and the published papers do
not offer detailed dataset descriptions and error analyses of the
content. Moreover, although receipts and invoices have some common
attributes, their analyses differ vastly due to complex graphical
layouts and richer content of the required pieces of information used
in invoices.
In their experiments with recently published Kleister
datasets~\cite{stanislawek2021kleister} that include financial
reports of charity organizations and non-disclosure agreements,
Stanis{\l}awek et al.\ found that results using state-of-the-art models
drop vastly with long, complex layouts in comparison to SROIE.
In 2006, Lewis et al.~\cite{10.1145/1148170.1148307} published the IIT
Complex Document Information Processing Test Collection (IIT-CDIP)
based on the Legacy Tobacco Documents Library, containing roughly
40~millions scanned pages for evaluation of document information
processing tasks.  In subsequent works, Harley et
al.~\cite{harley2015evaluation} took random samples of 16 categories
in IIT-CDIP, each with 25,000 images, for document image classification
evaluation. Nevertheless, none of those has benchmark results or gold
standard data for the information extraction task.

In this article, we present the latest development of the OCRMiner
project aiming at automated processing of (semi-)structured business
documents, such as contracts and invoices, based solely on the
analysis of OCR processing of the document pages.  The current system
significantly extends the original pipeline~\cite{ha2018recognition}
for information extraction.  We detail the improvements in each module,
i.e.\ more precise page segmentation as well as Long short-term
memory (LSTM) based OCR engine in text recognition, coping with OCR
errors in keywords using an approximate string matching algorithm, or
recognizing named entities using BERT based models. The presented
approach boosted the results for the English invoice dataset and
adapted well for Czech invoices.

The OCRMiner system
represents the documents as a graph of hierarchical text blocks with
automatic modular feature annotations based on keywords, text
structures, named entity processing and location parser.  With these
features, a rule-based method is built to extract the desired
information. The core idea behind OCRMiner lies in the human-like
approach to the document content analysis. Different from other works
overviewed in the next section, OCRMiner employs rich annotation from
several text analysis modules in combination with text positional
attributes. It does not only consider the absolute position of the text in the page
presented by the bounding box, but also
the relation with other groups of text.
OCRMiner is also able to cope with OCR errors using
a similarity search keyword matching algorithm. The system is unique
in distinguishing both local (company name, address, ...) and global
information such as buyer/seller identification. The higher-level
modules of OCRMiner (invoice page classification, block type
detection, ...) are language independent and the system is thus
portable to new languages.

The next section brings a detailed overview of the latest results in
invoice information extraction. Section~\ref{sec:ocrminer} details the
insides of OCRMiner and its modules. In Section~\ref{sec:eval}, we offer
a detailed evaluation of the system based on two datasets containing
invoices in the English and Czech languages from tens of different
suppliers.

\section{Related works}

Previous works aiming at automated processing of OCR
invoices,\footnote{We denote as ``OCR invoices'' all kinds of invoice
documents processed as an image by an Optical Character Recognition
tool. This includes both scanned paper documents as well as digital
born documents such as PDF.}
rely on predefined sets of layout of text data associated with
a specific invoice format.  One of the first
systems~\cite{bayer1997generic} used a specific programming language
for the description of the frame representation language for
structured documents (FRESCO).  For both text blocks and tables,
keywords are located first, then the data items are extracted
according to the programmed rules.  The system did not deal with other
items without a unique syntax. The recognition rate was between
40--60\%.

Adaptive rule-based systems that are based on layout information and
text keywords usually divide the rules into the \emph{vendor
dependent} and \emph{vendor independent}
knowledge~\cite{cesarini1997conceptual,cesarini2003analysis}. Cesarini
et al.\ in~\cite{cesarini2003analysis} introduces a multi-class
invoice analysis system (IAS) that interconnects the \emph{layout
structure} with the \emph{logical structure} of the analysed document
based on rules divided into the above mentioned two levels of
knowledge. A comparable approach is presented in the German system
named SmartFix~\cite{klein2004smartfix,schulz2009seizing}.  In this
system, the invoices are classified into known and unknown classes
based on layout graph using case-base reasoning technique (similar
to~\cite{hamza2007case}). 
Those systems need a high amount of annotated graph samples 
to reach an acceptable result.

A template-based approach is proposed
in~\cite{bart2010information,esser2014few}.
In~\cite{bart2010information}, users have to provide one entry of
a table or a list in a document, then the system tries to find
repeated structures which have the same or similar format. These
candidates are evaluated by the overall match quality in term of
perceptual coherence features such as alignment, height, width,
separation and gaps.  In~\cite{esser2014few}, their objectives are
towards small companies or single practitioners who cannot afford
a large training set nor 
a customized system.  The key idea is grouping documents generated by
the same template, then generating extraction rules for each group
based on at least a minimal number of similar training examples. The
final result is a combination of results from different indexers for
fields with a fixed value (e.g.\ the sender), a fixed position (e.g.\
the invoice number), a variant position and value (e.g. the total)
etc. User feedback is allowed to add annotated documents as a new
training example. For at least one training document per template, the
system reaches 78\% F1 score.

One of the main problems of processing invoices is the huge
variety of invoice layout formats used in practice. Although a known
set of metadata (i.e.\ the invoice number, date, the total, seller and
buyer) is obligatory in most cases, their layout and positions are
unconstrained.  Therefore, many invoice analysis systems consider
a particular layout as a class, usually connected to a particular
vendor. If an invoice is assigned the correct class, then this will
boost the extraction performance.  However, the system must manage
a large number of classes which may grow over time due to new
suppliers or to layout updates. Trying to avoid classifying invoice
into known and unknown classes which could lead to errors by
misclassification, Aslan et al.~\cite{aslan2016part} apply the
part-based modeling (PBM) approach from computer vision to invoice
processing based on deformable compositions of invoice parts
candidates obtained from machine learning based classification
techniques. The presented evaluation of invoice block detection ranges
from 69\% to 87\% with the average of 71\% accuracy.

Applying the results
from the deep learning area, the CloudScan
system~\cite{palm2017cloudscan} uses recurrent neural networks
(specifically the bidirectional long short-term memory
network~\cite{hochreiter1997long}) to learn the item dependencies
based on end-user provided feedback. The system reaches an F1 score of
84--89\% when trained on more than 300,000 invoices. A follow-up
system, named InvoiceNet~\cite{palm2019attend}, introduced a specific
deep network architecture denoted as Attend, Copy, Parse. The
architecture does not use word-level labels for the end-to-end
information extraction task, but it relies on experimentally designed
set of dilated convolution blocks for each of the operations. The
results improved the CloudScan results for four of the seven evaluated
item types.

Riba at al.~\cite{riba2019table} propose a method using Graph Neural
Networks (GNNs) for tabular layout detection based on the
classification of graph node embeddings. The method does not rely on
the document content although a commercial OCR tool is used for
encoding classifying entity textual attributes into numeric,
alphabetic, or symbol. The feature vector for each detected entity
(node of the graph) includes its bounding box position and a histogram
of textual attributes. An edge is created between two nodes if and
only if a straight horizontal or vertical line exists between bounding
boxes of the two nodes without crossing any other, excluding long
edges covering more than a quarter of the page. The best accuracy of
classifying 8~regions including address, table, or header lies between
82.7\% and 84.5\%.

Regarding the task of scanned invoice classification, in~\cite{tarawneh2019invoice} the authors proposed a method to
classify invoices into three categories: handwritten, machine-printed
and receipts, for the best selection of OCR approach in latter
processing. The features are extracted using a pre-trained Deep
Convolution Neural Network (CNN) Alexnet. The experiments with 1,380
documents showed that the K-nearest neighbors method achieved the best
result when compared to Random Forest (RF) and Naive Bayes (NB) with
98.4\% total accuracy. In a similar task, Jadli and
Hain~\cite{jadli2020automatic} test different combinations of
pre-trained CNN models (including VGGNet, Inception V3, DenseNet, and
MobileNet-V2), dimensional reduction techniques (Principal Components
Analysis, Linear Discriminant Analysis), and classification algorithms
with a small dataset of 200 samples involving electronic invoices,
hand written invoices, bank checks and receipts. The best result is
96.1\% using full features extracted using VGG19 and Logistic
Regression classifier. Similar techniques are used for document image
source classification with the results ranging from 93\% to 98\%
depending on the dataset
granularity~\cite{sharad2019first,tsai2019deep}.

CUTIE (Convolution Universal Text Information Extractor)~\cite{zhao2019cutie} applied
convolution neural networks to texts converted to semantic embeddings
in a grid-like structure.
In this work, position features are taken into account, however, the exact positional relations are only implied.
In~\cite{xu2020layoutlm,xu2020layoutlmv2}, the authors proposed LayoutLM(v2)
for pre-training text, layout, and image.
The idea is inspired by Bidirectional Encoder Representations from Transformers (BERT)~\cite{devlin2018bert}.
Although the image feature is used and the model works well on downstream tasks 
such as form understanding~\cite{jaume2019funsd}, receipt understanding~\cite{huang2019icdar2019}, 
or document image classification~\cite{harley2015evaluation},
same as in CUTIE system, the potential relationship between text segments
has not been taken into consideration. In addition,
sufficient data and time are required to pre-train the model.
A very similar idea on using text along with its position 
based on transformer architecture,
particularly BERT, is presented in~\cite{garncarek2020lambert}.
These systems need large datasets for pre-training models,
and smaller labelled datasets for fine-tuning. Such datasets
are not available by now for non-mainstream languages 
like the Czech language.

Experiments by Hamdi et al~\cite{hamdi2021information} with invoice
information extraction of document type and number, dates, amounts,
and currency show that word classification approach (similar
to~\cite{palm2017cloudscan}) outperforms sequence labelling
methods~\cite{garncarek2020lambert} on most fields.

Recently, graph-based models have been proposed
either using Graph Convolutional Networks~\cite{lohani2018invoice,yu2020pick}
or Graph Attention Networks~\cite{liu2019graph,krieger2021information} 
for sequence labelling
or node classification to identify the key items.
The graph-based document representations include positional relations between 
text segments but they are limited to the direct 
neighbours only. Furthermore, training the graph embeddings
needs a large amount of manually labelled document graphs. 

\begin{figure}[t]
\centering
\includegraphics[height = 0.5\textheight]{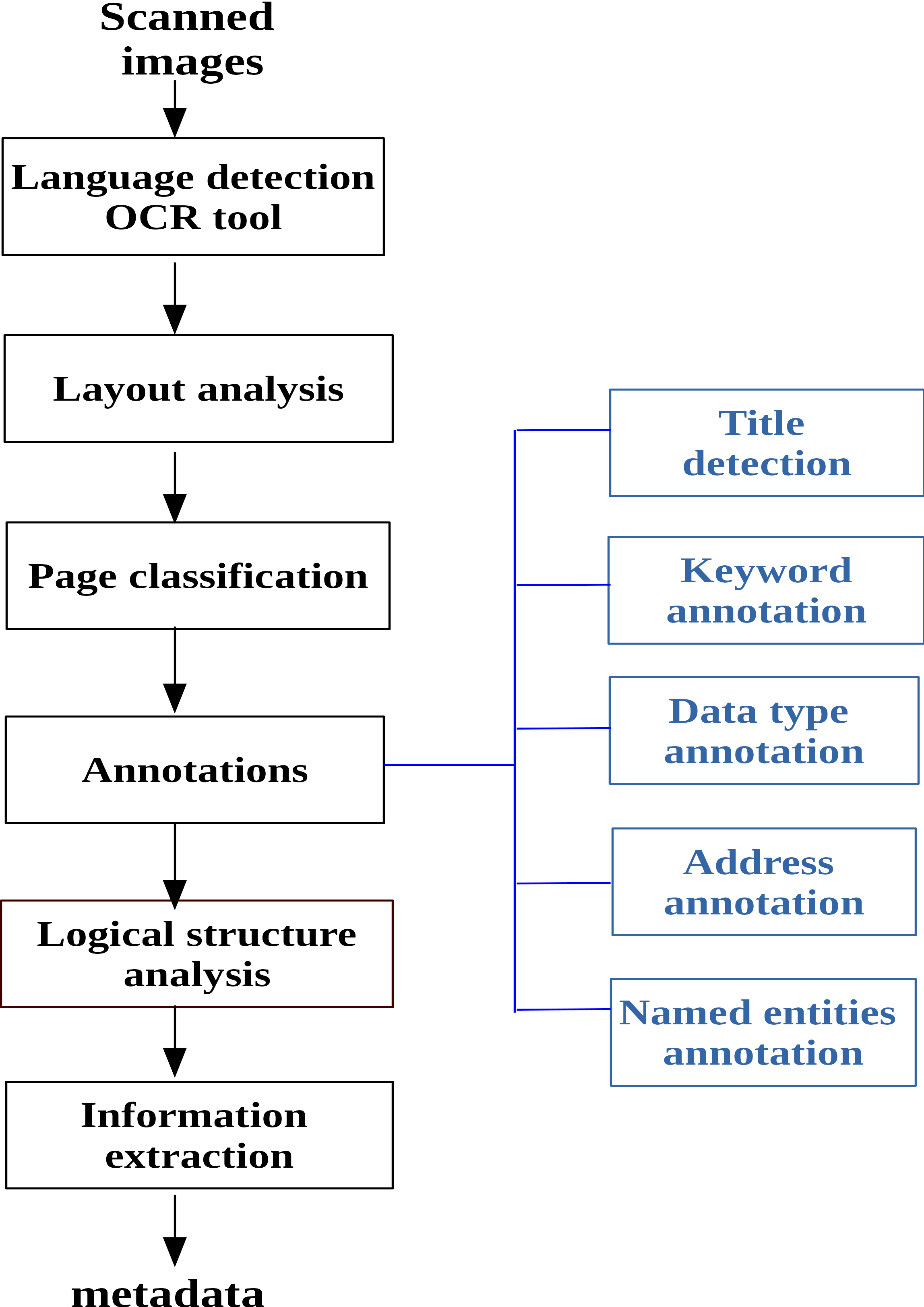}%
\caption{The processing pipeline}
\label{fig:pipe}
\end{figure}

A similar idea to the OCRMiner method that combines information 
from multiple sources of the data types and textual
content in relative positions is found in~\cite{majumder2020representation}.
In that system, for each field to be extracted, 
the generated candidates (e.g.\ dates, codes or numbers)
are scored by their neighbourhood word and position embeddings.
In contrast, OCRMiner begins with key phrases which drive 
the search for the data types in the visual surroundings 
inspired by the human-like reading of complex layouts.
A disadvantage of the other method lies in limiting 
the neighbourhood to a fixed window.
Since the layouts of invoices vary a lot, this approach 
can create unnecessary neighbours if the text is dense, 
and miss neighbours where the keyword and data are 
on the same line but distant. Moreover, OCRMiner not only 
deals with key texts matching fixed patterns, 
i.e.\ (invoice-, due-) dates or (invoice-, order-) numbers, 
but identifies also fields such as company names, contact
persons, or general addresses.

\section{The OCRMiner pipeline}
\label{sec:ocrminer}

A natural approach that humans use to extract
specific information from a document involves
2 steps: scanning the document to locate the
position of the information and then extract it
in the desired format. Some types of information
have unique structure defined by human rules, e.g.\ a VAT number.
Other types need context to differentiate them from the same or other types of
information, for instance, a date is distinctly identified as the invoice date or the due
date by the keyword which goes with it. And others may need heuristic knowledge,
for example the seller's name or address. The OCRMiner pipeline is
generally inspired by this human approach.

An overall schema of the OCRMiner processing pipeline is illustrated
in Figure~\ref{fig:pipe}.  The system is made of interconnected
modules that allow to add any kind of partial annotations to the
analysed document. First, the invoice image is processed by an OCR
tool.$\!$\footnote{See~\cite{ha2017recognition} for
details of the OCR setup.} When OCRMiner
needs to deal with  multilingual documents,
detecting languages used in the document is essential, not only for
setting up OCR tool, but also for latter annotations. The
main language of a document is defined as the language used by the title and data
field names. For instance, in invoices, these are ``invoice'', ``date'',
``order number'', ``payment method'', etc. Therefore, given a dictionary
of those terms for each possible language, if the document contains
the terms of a specific language, then this language is assigned as
the main language of the document. Otherwise, the language of the
document is defined based on text language distribution detected by the OCR
tool in the first run with a multiple languages setup.

Then, the physical document layout is built
based on the word and character positional information from OCR
(Layout analysis module). The Page classification\footnote{The
OCRMiner page classification is described in detail in
Section~\ref{sec:first_page}.} is performed with the information
obtained by this step in the pipeline.
From
that point, a series of annotations using different natural language
processing (NLP) techniques is applied to the content in the form of
XML tagging, involving title, keywords, data types, addresses and
named entities.  These annotations provide semantic information and
along with positional information, they create the context for further
analysis.
Based on these annotations, the logical structure, in
which each text block is assigned an informational type, is extracted
using a rule-based model. The final module extracts the desired
metadata information.

\begin{table*}[t]
    \centering
    \caption{Evaluation of the invoice first page detection module.}
    \label{tab:first_page}
    \begin{tabular}{r|c|c|c|c|c}
        Model & freq\_words & title+& freq\_words+ & annotation+  & all  \\
               &    &page\#       &title+page\#             & title+page\#  &features  \\
         \hline
         Naive Bayes Classifier:& & & & \\
         Precision & 93\% & 89\% & 95\% & 93\% & 95\%\\
         Recall & 94\%	&96\%	&96\%	&\textbf{99\%}	&98\%\\
         F1 score & 94\%	&92\%	&95\%	&96\%	&96\%\\
         \hline
         Logistic Regression:& & & & \\
         Precision & 97\%	&95\%	&\textbf{98\%}	&95\%	&97\%\\
         Recall & 97\%	&96\%	&97\%	&97\%	&98\%\\
         F1 score &97\%	&95\%	&\textbf{98\%}	&96\%	&\textbf{98\%}\\
         \hline
         Linear SVC:& & & & \\
         Precision &96\%	&96\%	&93\%	&90\%	&94\%\\
         Recall & 96\%	&96\%	&87\%	&85\%	&92\%\\
         F1 score &96\%	&96\%	&88\%	&86\%	&93\%\\
         
    \end{tabular}
\end{table*}

\subsection{Document physical layout analysis}

The analysis of document structure can be divided into the
\emph{layout structure} (also called physical structure or geometric
layout structure) and the \emph{logical structure}.
The layout structure analysis (or page segmentation) refers to the
process of segmenting a document page into groups of text lines, text
blocks, and/or graphical elements.  In this paper, the physical layout
structure is built using a bottom-up approach to obtain the document
layout structure from the output of an OCR tool.

The invoice image is first processed by the OCR engine\footnote{The
open source Tesseract-OCR~\cite{smith2007overview} version 4.1.0 is used now.},
which computes the positions (bounding boxes) of each recognized word.
Within the analysis, the words are grouped into lines based on three
criteria: alignment, style, and distance. If two words have similar 
alignment and style, and the distance
between them is less than a  threshold then they are in the same line.
The threshold was derived from the histogram of distances between each
couple of adjacent words in 215 invoice images
and currently corresponds to the height of the first word in the line.

The identified lines form (tabular) text blocks 
in a similar process, i.e.\ if two lines are 
vertically aligned, they have nearly the same font size, 
and the distance between them is less than a threshold,
then they are grouped into a block.
However, while the distances between words in a line usually
correspond to the space character, the distances between lines in
a block vary a lot depending on the graphical format. 
The best block-line threshold was set as twice as the height of
previous line in the block.

After deciding the document physical layout, each block receives an
identification of its named positions in the form of block attributes,
including the absolute position in the page and the relative position
of the block to other blocks. The former divides a page into 
header, top, middle, bottom, and footer in the 
vertical direction, and left, right parts in the 
horizontal direction whereas the latter searches 
for the block's neighbors in the top, bottom, 
left, right and bottom-right positions respectively.

\subsection{Page classification}
\label{sec:first_page}

\begin{figure*}
    \centering
    \includegraphics[width=.85\textwidth]{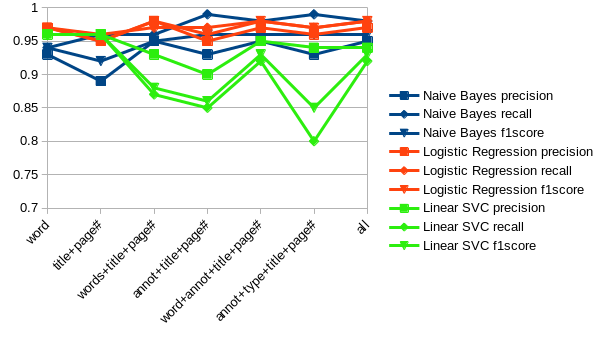}
    \caption{Evaluation of invoice first page classifiers using
    different features.}
    \label{fig:first_page}
\end{figure*}

The OCRMiner pipeline includes two classification tasks to answer the
questions whether the document is actually an invoice or not and if
the page is the first page. The reason is that business documents such
as invoices are often scanned and processed in batches.  OCRMiner thus
contains the detection of an invoice first page. Similar
to~\cite{wu2018visual}, the classification uses the available pieces
of the semantic content of the document. The extent of the set of the
available features depends on the module's position in the pipeline.
Usually, this classification step goes between the layout analysis and
the logical structure analysis. At this position, 
the features include binary features for the most frequent
words obtained from the document collection for each
language excluding entity names and stop words,$\!$\footnote{In the
experiment, these binary features (True/False meaning the word
appears/does not appear in the text)  are created for 111 words that
resulted from choosing 150 most frequent words and eliminating entity
names and stop words.}
the (identified) page title
(text, its position in the page and its size), and the (identified)
page number. In case when the classification module goes further in
the pipeline, the features are enriched with keywords and data type
annotations as well as block types. While the \emph{frequent words}
feature considers only individual words, the annotations bring the
logical meaning information.  As an example, a part of the feature
vector from an invoice first page looks as follows:\footnote{The
prefix \texttt{u""} stands for a \emph{frequent word}, \texttt{"K\_"}
for a \emph{keyword} annotation, \texttt{"D\_"} for a \emph{data type}
annotation. The presented frequent words are ``dodavatel -- seller'',
``odběratel -- buyer'', ``faktura -- invoice'', ``doklad --
document'', ``splatnosti -- due date'', ``zdanitelného -- taxable''.}
\begin{quote}

    \begin{tabbing}
        \small\tt \{ \= \kill
        \small\tt \{ \>\+ \small\tt u"dodavatel":~True, u"odběratel":~True,\\
        \small\tt u"faktura":~False, u"doklad":~True, \\
        \small\tt u"bankovní":~True, u"splatnosti":~True, \\
        \small\tt u"zdanitelného":~False, \ldots{}, \\
        \small\tt "title":~True, "top": 292, "height": 61, \\
        \small\tt "page\#": 1, "K\_BUYER":~True, \\
        \small\tt "K\_SELLER":~True, "K\_INVOICE NUMBER":~True, \\
        \small\tt "K\_PAY\-MENT METHOD":~True, \+\\
        \small\tt "K\_ORDER NUMBER":~False, \ldots{}, \\
        \small\tt "D\_ORGANIZATION":~True, "D\_DATE":~True, \\
        \small\tt "general info":~True, "buyer info":~True, \\
        \small\tt "seller info":~True~\}
    \end{tabbing}

\end{quote}
The classification output label is 1~if the page is a first page of an
invoice, and 0~for others. The latter means the page can be second or
further page of an invoice or not an invoice at all.
The first page classification module was evaluated with a Czech
dataset containing 1,150 pages (650 first invoice pages and 500 other
pages\footnote{The data set is a bit different from the one used
in~\cite{ha2017recognition}. In the current experiment, pages which
are not in Czech and extremely bad quality scans were removed and new
data was added.}) using 10-fold cross validation. 
The experiments employed several fast classifiers,
see Table~\ref{tab:first_page} for the average 
precision, recall and F1 score of Support
Vector Machines (Linear SVC),  Logistic Regression, and Naive Bayes.
The effect of the non-word features is clear in the group of SVC
classifiers whereas the two other groups are not affected to such
extent. Surprisingly, just the identified title and the page number
can serve for 96\% precision and recall. Moreover, by adding the
annotations only (without the \emph{frequent words}), the module can
recognize up to 99\% of the invoice first page.  Among the
classifiers, Naive Bayes reaches the highest recall of 99\% while
the Logistic Regression attains the best balance
between the precision and the recall with the highest F1 score of
\textbf{98\%} (this is in accordance with the results
of~\cite{jadli2020automatic}).
Other combinations of features and the visualization of the result are
illustrated in Figure~\ref{fig:first_page}.

\subsection{The annotation modules}

The annotation modules process the enriched OCR document and
supplement its content with valuable pieces of information based on
the output of natural language processing tasks. This part corresponds
to the human reader procedure, who identifies e.g.\ a name of a city
in various parts of the document and uses this information in the
global-level rules.

The first three modules in the series of task-oriented annotators operate
over plain text of recognised block lines to identify specific
keywords and structured data such as dates, prices, VAT number, and
IBAN. 
Keywords include words or phrases which are the signposts 
to find the desired piece of information. For instance, 
keywords for the invoice date are ``invoice date'', 
``date issued'', ``issued date'', ``billed on'', or 
``date'' at the beginning of a line. The set of 
keywords (per language) is obtained by analysing 
invoices in the development set. These modules 
can also cope with some OCR character-level errors.

The subsequent modules provide higher-level information such as named
entities (personal names, organizations, locations) and formatted
address specifications.

\begin{figure*}[t]\hfuzz=8pt
\begin{verbatim}
    seller info -> block_annot.data in [ORGANIZATION, PERSON, LOCATION, CITY, 
            COUNTRY, EMAIL, PHONE] and SELLER in top_blocks.block_annot.keyword 
            and top_blocks.num_lines == 1
\end{verbatim}
\caption{Block type rule example: if annotations in the block contains
more than at least one label in the set {\tt \{ORGANIZATION, PERSON, LOCATION, CITY, 
COUNTRY, EMAIL, PHONE\}} and the top block has only one line marking
the SELLER keyword, then the block type is {\tt seller info}.}

\label{fig:block_types}
\end{figure*}

\subsubsection{Named entity recognition}

The invoice content analysis relies heavily on the ability to identify
the relevant entities (typically person name, place name,
organization, or a product name, artwork, date, time) mentioned
in the tabular formatted texts.
The task of named entity recognition (NER) consists of two steps:
named entity identification (including named entity boundaries
detection in case of multi-word NER) and the entity classification.
OCRMiner currently employs pre-trained BERT NER model~\cite{bertlarge} 
based on the Google BERT-LARGE model~\cite{devlin2018bert} for English
and the SLAVIC-BERT model~\cite{arkhipov2019tuning} for the Czech language.

Although current state-of-the-art results in NER look promising for
both English and Czech, the text style of invoices causes the general
models to miss important context properties. First of all, the text
chunks are rather short in invoice blocks when compared to the full
sentences in existing models. The next challenge is that uppercase
formatting (which is an important feature in the NER models for Czech
and English) is used more frequently in this type of documents, for
example, for headings or company names.  The third problem is strong
multilingual characteristics of many invoices, e.g.\ in English
invoices, the names of organizations or streets can be in a different
(local) language.

\subsubsection{Location names recognition}

Beside the location entities detected by NER, OCRMiner uses a global
address parser \texttt{libpostal}~\cite{libpostal2020}, a statistical
model based on conditional random fields trained on Open Street
Maps\footnote{\url{http://openstreetmap.org/}} and Open
Addresses.$\!$\footnote{\url{http://openaddresses.io/}} The reasons
for this doubling reflect the importance of addresses in the invoice
metadata and the need to be able to solve possible conflicts in the
name classification.  Each of the two modules has positive matches on
different data.  For example, street names such as
\emph{Mahenova 9/181}
are well recognized by \texttt{libpostal}
while the NER module recognizes \emph{Mahenova} as
person name. On the other hand, the location names recognition matches
\emph{Konica Minolta Business} as an office building in the U.S. while
the NER module annotates it (correctly) as an organization name.

\subsection{Block type detection}
\label{sec:block_type}

After the annotation modules add all relevant local pieces of
information to the analyzed content, each block is assigned 
possible block types.  The invoice text blocks are categorized into:
\begin{itemize}
\item general information containing invoice number, order number,
    date;
\item seller/buyer information including company name, address, vat
    number, contact detail;
\item delivery information such as delivery address, date, method,
    code and cost;
\item bank information covering the bank name, address, swift code,
    account number;
\item the invoice title and page number.
\end{itemize}
Blocks that do not belong to any of these categories are assigned an
empty label.

The OCRMiner block type detection technique is based on a set of
logical rules that combine information obtained in the preceding pipeline
steps. The rules are in human readable and easy to edit form (see an
example in Figure~\ref{fig:block_types}). Each rule is applied to each
block in the invoice document. If a block meets the rule's condition,
then the label is added to the block type.
A detailed evaluation of this module with a prototype English dataset
is provided in~\cite{ha2018recognition}.

\subsection{Information extraction}

\begin{figure*}[t]
    \centering
    \includegraphics[width = .6\textwidth]{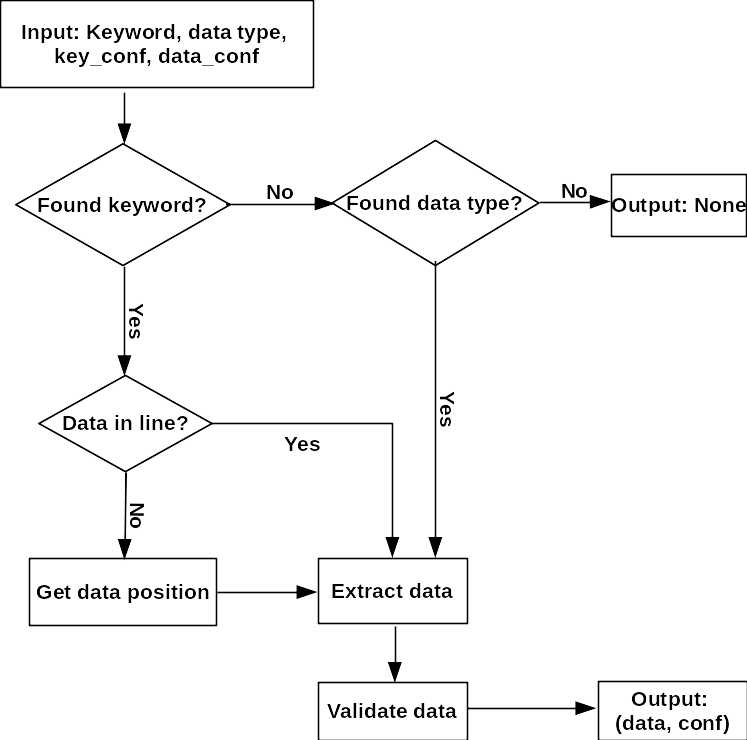}
    \caption{The information extraction workflow process.}
    \label{fig:ext_diagram}
\end{figure*}

The extracted data can be a single piece of information such as the
invoice number, invoice date, the order number, the order date, the
due date, the payment date, the payment method, IBAN, or the swift
code, or a group of information such as the buyer information, seller
information and the delivery address. The first group can be divided
into smaller categories: ``DATE'', ``PRICE'', ``NUMBER'', and
``GENERAL''. The first two clusters usually attach keywords with data
types whereas the two others often have no rule for the data. Unique
structured information such as VAT number can be revealed without
keywords.

Each type of information has a confidence score attached
distinguishing the situations when only the keyword was found, the
data type matches, or both.

\filbreak
\noindent
For instance, with the VAT number as an
example:
\begin{verbatim}
    {"key": VAT NUMBER,
     "data": VAT NUMBER,
     "key_conf": 0.7, 
     "data_conf": 0.8,
     "combine_conf": 1.0}
\end{verbatim}
If both the keyword and the data type are found, then the confidence
score is set to 1.0. Otherwise, if only the keyword or the data type
is found for some reason (e.g.\ OCR errors), then the confidence is
lowered to 0.7, or 0.8 respectively. The confidence score values are
based on the development set evaluation.

The extraction workflow process is illustrated in
Figure~\ref{fig:ext_diagram}.  First, the system searches for the
appropriate keywords in the annotations. This returns the block and the
line containing the keyword if the keyword is found. Then, if the line
does not contain the expected data, a weighted function is applied to
the right block and bottom/bottom-right block to see 
if the data lies on the right of the keyword or under the keyword.
This function sets a high score to the correct data type, a lower score to
any other data found in the annotations, and a penalty for other
keywords founds. The reason is that the information is often organized in
columns where keywords are in the left column and data on its right.
The higher weighted block is more likely to contain the data. If the
keyword is not found, then the system will look at the data type in
case it is structured information. The annotation steps are able to
cope with some OCR errors, so, the validation step checks and corrects
the data. For example, the Czech VAT number begins with ``\texttt{CZ}''
and not ``\texttt{C2}'' or the email symbol is ``\texttt{@}'' and not
``©''.

\begin{figure}[b]
\centering
\begin{minipage}{.90\columnwidth}
\begin{verbatim}
If block is vertically aligned or horizontally
    aligned with SELLER block and
    has no different content in [COMPANY,
        ADDRESS, ID, VAT NUMBER] 
    compared to SELLER, 
then label is SELLER.
\end{verbatim}
\end{minipage}
\caption{A buyer/seller classification rule example}
\label{fig:selbuy_rule}
\end{figure}

\begin{figure*}[t]
    \centering
    \begin{tabular}{ll}
         ~~a) & ~~c)  \\
    \setlength{\fboxsep}{1pt}%
    \fbox{\includegraphics[width = .25\textwidth,valign=t]{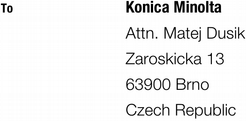}}%
    & \includegraphics[width = .69\textwidth,valign=t]{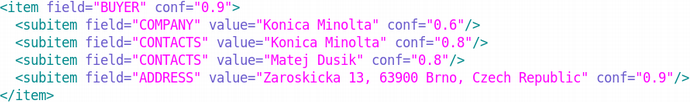}\\
    ~~b)&\\
    \multicolumn{2}{l}{\includegraphics[width = 0.97\textwidth]{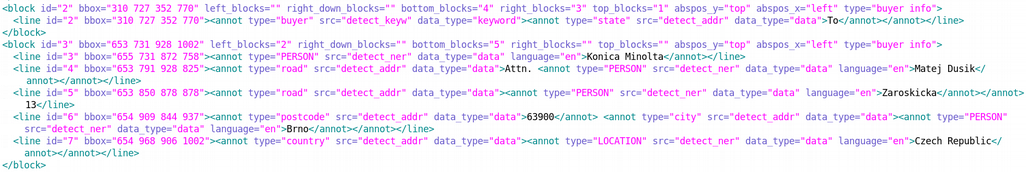}}
    \end{tabular}
    \caption{An example of an address block analysis: a) the input
        image, b) after annotation, and c) the extracted information.}
    \label{fig:pipeline_eg}
\end{figure*}

The blocks with buyer and seller information are not always
accompanied by keywords and their information may even be spread into
several blocks. For those blocks, the information is classified into
buyer's or seller's using a set of logical rules. These rules consider
the block position in the page (e.g.\ prevalently, the seller's
contact information is located at the bottom of the page), and
alignment and content comparison with the found seller/buyer blocks.
Finally, global rules based on the observation that the seller block
often comes before the buyer block in vertical or horizontal direction
are applied to blocks unprocessed in previous steps. Currently,
OCRMiner uses 15 classification rules and 5 global rules written in
descending confidence order. If a block satisfies a rule, then the
corresponding label is returned and the system ignores the remaining
rules.  A classification rule example can be found in
Figure~\ref{fig:selbuy_rule}.

Different from other types of information which are mostly expressed
in a text line, addresses usually spread in more than one line, in
particular cases even with other text in between. Therefore, the
address extraction does not simply extract an annotated
\emph{location} entity.  In this step, the Libpostal address parser is
employed with detailed address parts recognition.
The extraction step takes ``road'' (street name) and ``house\_number'' as
important signpost to start an address. The labels are sorted in the
order of ``country'', ``city'', ``city\_district'', ``suburb'', ``postcode'',
``road'', ``house\_number'', ``house''. The labels at the beginning of the
list have higher rank than the labels at the end.  Then, the text line
containing the highest ranked label is considered as the end of the
address. If the extracted address contains a label at the beginning of
the list, i.e.\ a country or a city then the confidence number is
higher than for the other labels (the address may miss some part of
information). A LOC entity annotation also increases the confidence
number. Example of an address block analysis is presented in
Figure~\ref{fig:pipeline_eg}.

\section{Experiments and evaluation}
\label{sec:eval}

\subsection{The invoice dataset}

\begin{figure*}
\centering
    \begin{tabular}{ll}
         \includegraphics[width=0.4\textwidth, valign=t]{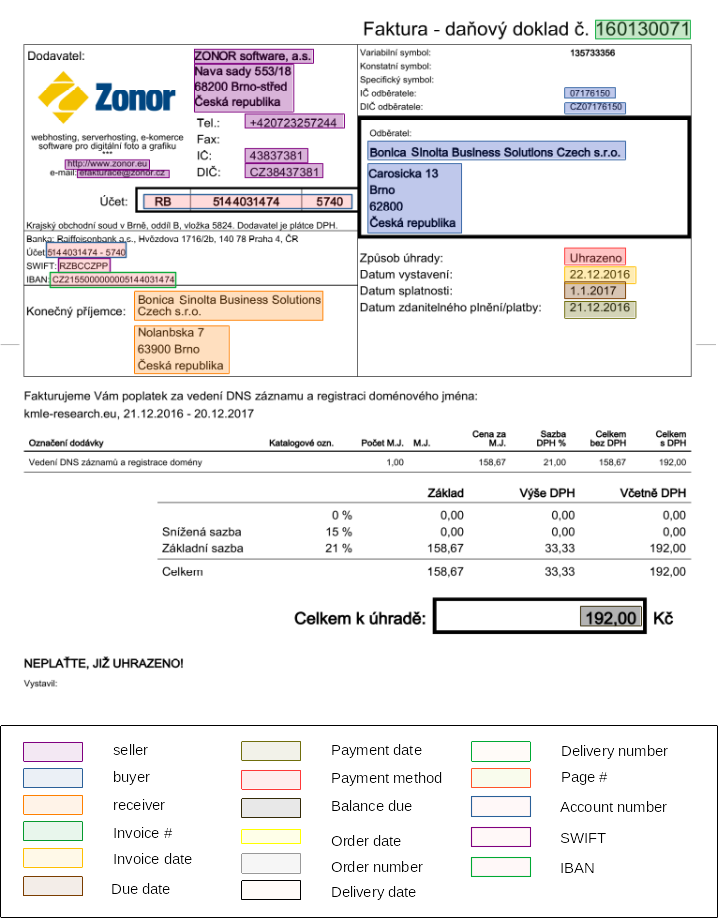}
         & \includegraphics[width=0.4\textwidth, valign=t]{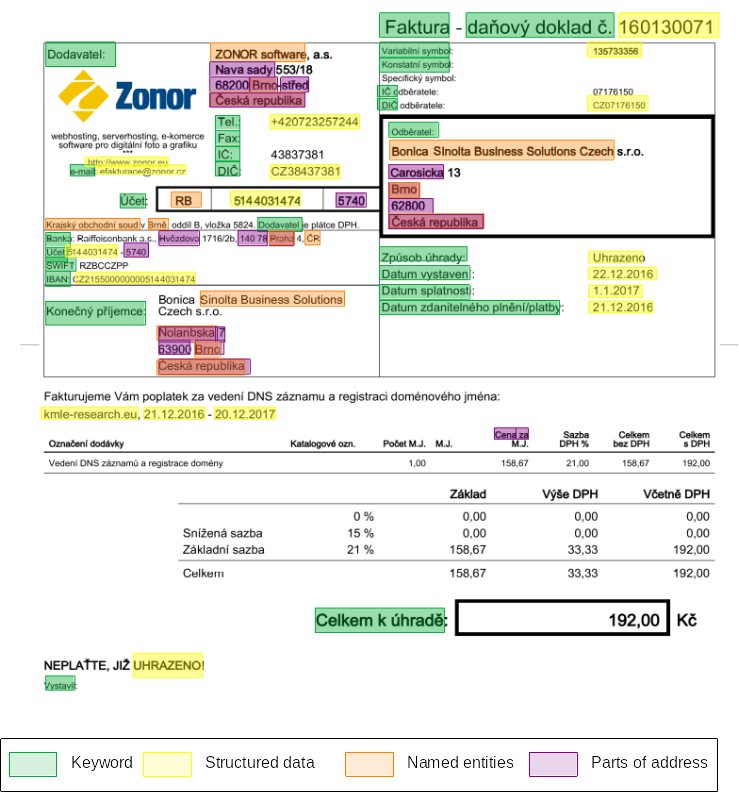}\\
         a) Invoice with the desired information
        & b) Invoice with annotations\\
    \end{tabular}
    \caption{The input invoice dataset examples}
    \label{fig:example}
\end{figure*}

The experimental invoice dataset is collected in cooperation with
a renowned copy machine producer. Due to the sensitiveness of the
invoice content, the dataset is not publicly available. The collection
consists of two main parts: English invoices and Czech invoices, see
Figure~\ref{fig:example}. The English group consists of invoices from
more than 50 suppliers all over the world whereas the Czech data comes
from over 100 vendors in the Czech republic.

In each data set, for developing and testing purposes 60 invoices were
randomly selected. Those 120 invoices are from 76 vendors. 
Each vendor has a different layout, even the layouts of 
invoices from the same vendor change over time.
Ten out of the 60 are used as a development set.
For instance, the ten English development invoices
are by nine different suppliers in
Austria, Poland, US, UK, the Netherlands, Germany, and Italy. More
than a half of the 50 testing invoices are from 18 other suppliers
which have not appeared in the development set. 
The annotated XML files expose detailed positional information of
7,602 text blocks,
18,672 lines, and 9,096 automated annotations of 64 annotation types.
The files have also been manually annotated for the metadata items
containing altogether 1,523 values of 24 types for the information
extraction process.
A detailed overview of the information items is displayed in
Table~\ref{tab:item_types}.

\subsection{Evaluation}

In the system evaluation,$\!$\footnote{The information extraction task
is evaluated independently from the page classification 
module which is fully described and evaluated in 
Section~\ref{sec:first_page}.} the extracted field values are classified
into 3 result classes. Each piece of information is evaluated as
a \emph{match}, a \emph{partial match}, or a \emph{mismatch}.  For
other data than company name and address, the \emph{match} means
always an exact match.  With company name and address, the
\emph{match} classification allows to ignore differences in
unimportant pieces of the gold standard values. A \emph{match} is
assigned if the extracted data contains the main information, even
though some words are missing or surplus. 
For example, a company name of:
\begin{quote}
    \begin{description}
    \item[ground truth:]~\\\raggedright
            \texttt{Konica Minolta Business Solution Czech spol, s.r.o.}
    \item[extracted:]~\\\raggedright
            \texttt{Konica Minolta Business Solution Czech spol, s.r.0}
    \end{description}
\end{quote}
are considered a \emph{match}. In this example,
the letter ``o'' in ``s.r.o.'' was recognized as ``0'' (zero) 
by the OCR engine. In this context, such OCR errors 
are easy to identify and fix as the ``tail'' of the 
name contains the known organization type (``s.r.o.'' 
in Czech corresponds to ``Ltd.'' in English).
Or the address:
\begin{quote}
    \begin{description}
    \item[ground truth:]~\\\raggedright
        \texttt{Level 6, 341 George St, Sydney NSW 2000, Australia}
    \item[extracted:]~\\\raggedright
        \texttt{Atlassian Pty Ltd, Level 6, 341 George St, Sydney NSW
        2000, Australia}
    \end{description}
\end{quote}
are also regarded as \emph{match}ing.

\begin{table*}[t]
    \centering
    \caption{The most common item types in the development and test sets} 
    \label{tab:item_types}
    \begin{tabular}{l|r|r|r|r}
         Type of info & En dev set & En test set & Cz dev set & Cz test set \\
         \hline
         Invoice number & 10~~ &  45~~ & 10~~ &  49~~ \\
           Order number &  6~~ &  12~~ &  6~~ &   8~~ \\
                  Swift &  6~~ &   8~~ &  6~~ &  20~~ \\
         Account number &  8~~ &   9~~ &  8~~ &  45~~ \\
            Page number &  4~~ &   7~~ &  4~~ &  16~~ \\
           Invoice date & 10~~ &  47~~ & 10~~ &  46~~ \\
               Due date &  7~~ &  23~~ &  8~~ &  45~~ \\
           Payment date &  8~~ &  10~~ &  8~~ &  37~~ \\
              Total due &  9~~ &  27~~ &  9~~ &  25~~ \\
                   IBAN &  7~~ &   9~~ &  7~~ &  22~~ \\
         Payment method &  8~~ &  18~~ &  8~~ &  34~~ \\
           Company name & 23~~ & 101~~ & 23~~ & 114~~ \\
               Contacts &  6~~ &  24~~ &  6~~ &  21~~ \\
                Address & 25~~ & 101~~ & 25~~ & 119~~ \\
             Vat number & 18~~ &  59~~ & 18~~ &  91~~ \\
                  Email &  8~~ &  24~~ &  8~~ &  35~~ \\
                Website &  3~~ &  25~~ &  3~~ &  23~~ \\
           Phone number & 15~~ &  17~~ & 15~~ &  40~~ \\
             Company id &  0~~ &   0~~ & 20~~ &  99~~ \\
            Amount paid &  0~~ &  21~~ &  0~~ &   0~~ \\
    \hline
    \end{tabular}
\end{table*}

\begin{table}[b]
\begin{center}
	\caption{Information extraction evaluation results for the English set}
	\label{tab:ex_eval_en}
	\begin{tabular}{ l|r|r|r|r|  }
            & \multicolumn{2}{c}{regular expressions} & \multicolumn{2}{|c|}{similarity search}\\
            \cline{2-5}
                          & match & in \% & match &        in \%   \\
            \hline
            Match         &523  &84.22   & 527 & \textbf{84.86} \\
            Partial match &35   & 5.64  &  35 &  \textbf{5.64} \\
            Mismatch      &63   & 10.14  &  59 & \textbf{9.50} \\
            \hline 
            Total         &621  &100.00  & 621 &        100.00  \\
	\end{tabular}
	
\end{center}
\end{table}

A \emph{partial match} is assigned if the extracted information
contains correct information, but some part is missing or some
character is incorrectly recognized by the OCR tool.  For instance, if
the OCR tool recognized ``\texttt{0}'' character as ``\texttt{O}'' in the
order number ``\texttt{BXOEF24CA4E2}'' instead of ``\texttt{BX0EF24CA4E2}'', or
an address misses the street name:
\begin{quote}
    \begin{description}
    \item[ground truth:]~\\\raggedright
        \texttt{Zarosicka 4395/13, Brno, Jihomoravsky kraj 62800,
        Czech Republic}
    \item[extracted:]~\\\raggedright
        \texttt{Brno, Jihomoravsky kraj 62800, Czech Republic}
    \end{description}
\end{quote}
The \emph{partial match} is ``correctable'' in the sense of finding
regular OCR character mismatches or via the possibility to search for
the expanded value in the identified block.

The comparison between the ground truth and the extracted information
in the evaluation is first preprocessed automatically using the
Levenshtein distance. The result returns a \emph{match} if the
distance is zero, a \emph{partial match} if the edit distance is less
than a threshold (set to~2 characters in the experiment) and
a \emph{mismatch} otherwise.  Then, due to the complexity of address
and company name blocks, the evaluation is double-checked manually.
For the group-type information such as seller and buyer, the result
counts are counted for each subfield, i.e.\ the company name, the
address, and so on.  This means that even if all information including
company name, address, company id, vat number are correctly extracted
but the group is wrongly classified as seller or buyer, then the
evaluation module will count all the subfields as mismatches.

\begin{table}[t]
\begin{center}
	\caption{Information extraction evaluation results with the Czech set}
	\label{tab:ex_eval_cs}
	\begin{tabular}{ l|r|r|r|r|  }
            & \multicolumn{2}{c}{regular expressions} & \multicolumn{2}{|c|}{similarity search} \\
            \cline{2-5}
                          & match & in \%  & match & in \%    \\
            \hline
            Match         &679	& 75.28    &   756 & \textbf{83.81} \\
            Partial match &30	& 3.33     &    40 & \textbf{4.43}  \\
            Mismatch      &193	& 21.40    &   106 & \textbf{11.75} \\
            \hline
            Total         &902    &100.00  &   902 & 100.00   \\
	\end{tabular}
\end{center}
\end{table}

As can be seen in the evaluation of the page classification module,
the keyword features are the most important in the correct
identification of the block types. To be able to cope with OCR
character misclassifications, OCRMiner contains two different modules
to search for keywords. The first module uses regular expressions
patterns and the second module is based on an approximate
similarity search. Within the similarity search, the text substrings
are evaluated with weighted edit distance 
(Levenshtein distance) measure which includes
a small cost (0.1 in the experiments) for a substitution of common
errors (such as `\texttt{l}', `\texttt{t}' and `\texttt{f}', or
`\texttt{u}' and `\texttt{v}') and insertion or deletion of
punctuation or white space, while any other edits have the cost of
1.0\footnote{See~\cite{ha2019approximate} for more details.}. 
The substring of the text line which has the weighted edit distance
less than a threshold (15\% of the length of the keyword) is marked as
the keyword.  The results of these two modules for the English and
Czech evaluation sets are presented in Tables~\ref{tab:ex_eval_en}
and~\ref{tab:ex_eval_cs}.  The similarity search module adapts better
to OCR errors, which leads to slight improvements in both the
languages when compared to the method using regular expressions match
-- 0.64\% or 3.22\% of mismatches are solved or partially solved in
the English and Czech datasets respectively.
In general, the whole system works better for the English set where
above 90\% of ground truth items are correctly identified (with exact
or partial match) compared to about 88\% of the Czech invoice items.

\begin{figure*}[t]
    \centering
    \begin{tabular}{ll}
    ~~a) & ~~b) \\
    \includegraphics[width = .47\textwidth,valign=t]{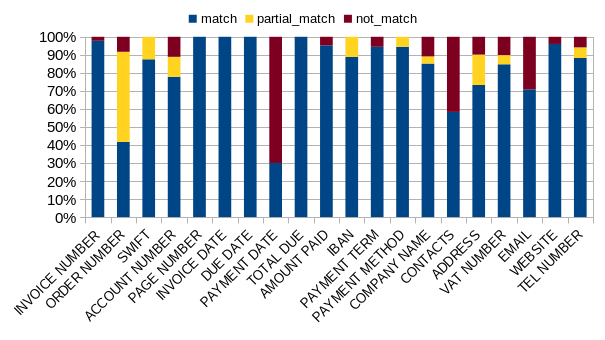}%
    & \includegraphics[width = .47\textwidth,valign=t]{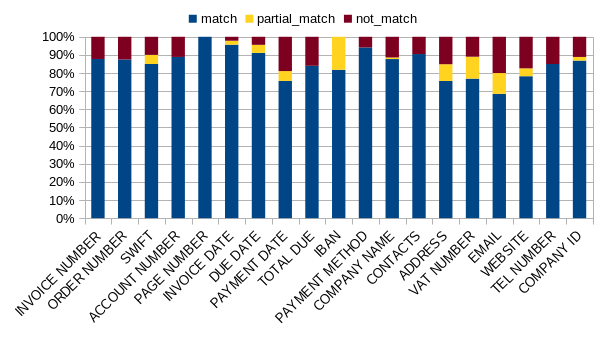}%
    \end{tabular}
    \caption{Evaluation of individual fields for a) the English
        invoices, and b) the Czech invoices.}
    \label{fig:w_his}
    \label{fig:ind_eval}
\end{figure*}

A detailed analysis of the extraction accuracy for each field is
illustrated in Figure~\ref{fig:ind_eval}. 
Both datasets have high match ratios at the page number, the invoice
date, the due date, and the payment method. The extraction works
better for the English invoices with items such as the invoice number,
the VAT number, IBAN, the total due amount, or the website. 
In contrast, 
the Czech dataset has better extraction results with the order
number, the account number, the payment date, and the contact
information.  There is an exceptional low match ratio for the payment
date in the English dataset due to OCR errors in a subset of English
invoices from a particular provider.
English invoices also witness a high partial match ratio 
with the order number. Different from Czech invoices, 
where the order number is a digital series, in the English dataset, 
the order number is often a mixture of numbers and letters, for example
``AP0CDF89CC10''. The first number ``0'' is between letters, 
and the OCR engine often misrecognises it as the letter ``O'', 
causing the high partial match.
Another high mismatch is the contact person item which
is caused by the situation when NER misrecognized a person as 
an organization due to capitalized letters.
A detailed error analysis is given in Section~\ref{sec:error_analysis}.

\hbadness=1500
For a performance comparison, the OCRMiner dataset has been evaluated
with the state-of-the-art InvoiceNet
system\footnote{\url{https://github.com/naiveHobo/InvoiceNet}} which is
based on the work of Palm et al.~\cite{palm2019attend}. InvoiceNet
solves the end-to-end information extraction task using deep neural
network models. 
Six common invoice fields were selected for the evaluation: the invoice number, the
invoice date, the seller's name and address and the buyer's name and
address. Each field requires a separate model. 
Since the original pre-trained
models are not publicly available, InvoiceNet has now been trained in
three possible setups. First, the InvoiceNet models have been trained
with the SROIE datasets~\cite{jaume2019funsd} including 727~receipts
(denoted as InvNet1). The second system (InvNet2) has been trained
with 208 English invoices of the OCRMiner dataset and the last one has
been trained with 500 Czech invoices from the same dataset (InvNet3).
InvNet1 and InvNet2 are evaluated with the English test set
whereas InvNet3 is evaluated with the Czech test set. The
results are summarised in Table~\ref{tab:invnet}.

\begin{table*}[t]
    \caption{Performance comparison on specific fields in \%}
    \label{tab:invnet}
    \centering
    \begin{tabular}{l|l|c|c|c||c|c|}
                         & & \multicolumn{3}{c||}{English data} & \multicolumn{2}{c|}{Czech data} \\
         field        & measure       & InvNet1 & InvNet2        & OCRMiner       & InvNet3 & OCRMiner \\
         \hline
         inv\_number  & match         & -       & 46.67          & \textbf{97.78} & 83.67   & \textbf{87.76}\\
                      & partial match & -       & 0.00           & 0.00           & 4.08    & 0.00\\
                      & mismatch      & -       & 53.33          & 2.22           & 12.24   & 12.24\\
         \hline
         inv\_date    & match         & 0.00    & 0.00           & \textbf{100.0} & 0.00    & \textbf{95.65}\\
                      & partial match & 0.00    & 0.00           & 0.00           & 0.00    & 2.17\\
                      & mismatch      & 100.0   & 100.0          & 0.00           & 100.0   & 2.17\\
         \hline
         seller\_name & match         & 12.24   & 44.90          & \textbf{83.67} & 77.27   & \textbf{79.55}\\
                      & partial match & 0.0     & 2.04           & 0.00           & 9.09    & 9.09\\
                      & mismatch      & 87.76   & 53.06          & 16.33          & 13.46   & 11.36\\
         \hline
         seller\_addr & match         & 4.08    & 6.25           & \textbf{81.63} & 2.00    & \textbf{86.00}\\
                      & partial match & 16.33   & 87.50          & 14.29          & 82.00   & 4.00\\
                      & mismatch      & 79.59   & 6.25           & 4.08           & 16.00   & 10.00\\
         \hline
         buyer\_name  & match         & -       & 71.11          & \textbf{86.67} & 82.00   & \textbf{84.00}\\
                      & partial match & -       & 2.22           & 8.89           & 8.00    & 2.00\\
                      & mismatch      & -       & 26.67          & 4.44           & 10.00   & 14.00\\
         \hline
         buyer\_addr  & match         & -       & 2.22           & \textbf{66.67} & 0.00    & \textbf{80.00}\\
                      & partial match & -       & \textbf{93.33} & 20.00          & 30.00   & 8.00\\
                      & mismatch      & -       & 4.44           & 13.33          & 70.00   & 12.00\\
    \end{tabular}
    
\end{table*}
Overall, OCRMiner performs better than In\-voice\-Net in all fields with
both the English and the Czech dataset. InvoiceNet locates the
position of addresses in English invoices better than OCRMiner but
fails to extract the full data. In multiple lines, multi-word text data
(e.g.\ an address or a company name) are identified only partially
with important parts missing. In most cases, only the street name and
house number are extracted.  The results also show that a system
trained with receipts cannot be generalised to invoices.
\begin{table}
    \caption{Evaluation of location detection by the combination of
    Libpostal and Named entity recognition (NER) modules.}
    \label{tab:loc_detect}
    \centering
    \begin{tabular}{ l|r|r|r|r }
        English set:        & \multicolumn{2}{l|}{Libpostal} & \multicolumn{2}{l}{NER} \\
        \hline
        both modules        &  287 & 25\% & 287 & 83\% \\
        one module only     &  876 & 75\% &  57 & 17\% \\
        ~--~true positives   &  230 & 20\% &  23 &  7\% \\
        ~--~false positives  &  646 & 55\% &  34 &  10\% \\
        \hline 
    \end{tabular}

        \smallskip
        \begin{tabular}{ l|r|r|r|r }
        Czech set:          & \multicolumn{2}{l|}{Libpostal} & \multicolumn{2}{l}{NER} \\
        \hline
        both modules        & 165 & 12\% & 165 & 59\% \\
        one module only     & 1223 & 88\% &  117 & 41\% \\
        ~--~true positives   & 292 & 21\% &  116 & 41\% \\
        ~--~false positives  & 931 & 67\% &  1 & 0\% \\

        \hline 
    \end{tabular}
\end{table}

\begin{table}[t]
\caption{Ablation study with the English dataset}
    \label{tab:ablation}
    \centering
    \begin{tabular}{l|c|c|c}
        Features         & match   & partial match & mismatch\\
        \hline
        full model       & 84.86\% & 5.64\%        & ~9.50\% \\
        without keywords & 61.06\% & 3.04\%        & 35.90\% \\
        without data     & 49.20\% & 1.12\%        & 49.68\% \\
        \hline
    \end{tabular}
\end{table}

\begin{table*}[t]
    \caption{Detailed analysis of the error types}
    \label{tab:error_types}
    \begin{center}
    \renewcommand{\tabcolsep}{4pt}
    \begin{minipage}[t]{.35\textwidth}
        \begin{tabular}{|r||r|r|r|r|}
        \hline
            \multicolumn{5}{|c|}{Mismatch} \\
            \hline
            error types        & \multicolumn{2}{c|}{En set} & \multicolumn{2}{c|}{Cz set}\\
            \hline
            Misclassified role &          27\% & 16          & \textbf{33\%} &  35\\
            OCR errors         &          10\% &  6          &          10\% &  11\\
            Misclassified NE   & \textbf{41\%} & 24          &           7\% &   7\\
            New keywords       &          12\% &  7          &          12\% &  13\\
            Layout             &           3\% &  2          &          17\% &  18\\
            Other              &           7\% &  4          &          21\% &  22\\
            \hline
            \textbf{Total}     &         100\% & 59          &         100\% & 106\\
            \hline
        \end{tabular}
    \end{minipage}\hspace{3em}%
    \begin{minipage}[t]{.3\textwidth}
        \begin{tabular}{|r||r|r|r|r|}
            \hline
            \multicolumn{5}{|c|}{Partial matches}\\
            \hline
             error types         & \multicolumn{2}{c|}{En set} & \multicolumn{2}{c|}{Cz set}\\
             \hline
             OCR errors     &          34\% & 12          & \textbf{66}\% & 29\\
             Addresses      & \textbf{46\%} & 16          &          14\% &  6\\
             Other          &          20\% &  7          &          20\% &  9\\
             \hline
             \textbf{Total} &         100\% & 35          &         100\% & 44\\
            \hline
        \end{tabular}
    \end{minipage}
    \end{center}
\end{table*} 

Table~\ref{tab:loc_detect} displays an evaluation of the location
detection by the two modules employed in the process -- Libpostal and
the named entity recognition (NER) module. 
Both modules mostly agree on city and country names.
Libpostal offers detailed annotation types of ``road'', 
``postcode'', ``suburb'', ``city\_district'', ``city'', 
``state district'', ``state'', and ``country'' labels. Of these, 
the postcode and street are the most important 
regarding the final scoring.
Consequently, Libpostal module has a higher recall 
than the NER model,
while the NER module offers better precision.
Libpostal is also not so sensitive to uppercasing, 
e.g.\ BRNO (a city), or CZECH REPUBLIC.
This is why an ensemble module is used in the system which allows to
take advantage of each model's capabilities. 
The system also requires a combined 
information support of at least two detected address items in the
analysed text block to eliminate most of false positive errors.

To better understand the contribution of keyword and data annotations
to the extraction, a set of ablation test were run for OCRMiner. In
each test, one of those two annotation types is left out and the
information extraction is based on the remaining annotations. The
extraction process can still identify some fields but with a decreased
value of the confidence score.
The overall results evaluated with the English test set are shown in
Table~\ref{tab:ablation}.  As expected, when keywords are missing, the
match ratio has fallen off for fields relying on the keyword only,
such as the invoice number, the SWIFT code, or the account number with
the drop of 97.78\%, 87.75\%, and 77.78\% respectively. The textual
fields such as the company name or address and the contact person are
slightly affected with the drop of 0.99\%, 4.16\%, and 6.93\%
respectively. In contrast, these fields' \emph{match} ratios drop
quickly (by 62.38\%, 58.33\%, and 73.27\%) when the data annotations
are absent.

\subsection{Error analysis}
\label{sec:error_analysis}

A detailed analysis of errors in the system output was evaluated based
on the results of the similarity search module. The causes of errors can
be categorized into five groups, but their ratios in each dataset are not
the same.

Different from other pieces of extracted information, the seller/buyer/delivery
addresses are groups of information. They are 
often not denoted by specific keywords, fragmented into several
blocks and their exact distinction is left to the reader.
To extract this information, the system needs to classify the
specified blocks into the seller, the buyer and the delivery address
based on their position and content.  
27\% of errors in the English invoices and 35\% of errors in
the Czech ones
are correctly extracted but misclassified in these three categories.
The combination of reasons for this type of errors includes errors in
detecting the addresses and named entities (false positives),
insufficient rule details, and an extra variety in the layout formats
(e.g.\ the position relation between the seller and the buyer without
a \emph{keyword} distinction) that makes the global rules conflicting
in the difficult cases.

While the misclassification causes the largest portion of errors 
in the Czech invoices (33\%), incorrect identification of entities 
forms the main reason of mismatches in the English invoices
(41\% of errors) including persons classified as 
organizations (``PETR GOTTHARD'', ``EVA LYSONKA'',...)
or undetected organizations and locations.

OCR errors and unknown keywords cause almost
the same portion of errors in each dataset with 10\% and 12\% respectively.
The Czech invoices display a broad range of layout formats which is
projected to 17\% of layout errors compared to only 3\% in the English set.
Most of these errors happen with special cases when the keyword and the data
are in an unexpected relative position, e.g.\ the keyword below the
data, or one line of a company info in a distant stand-alone block.

The error analysis of partial matches revealed that
the OCR errors cause 64\% of errors for English, and 34\% for the Czech invoice items.
A significant portion of these errors happens
in email addresses where the ``at'' symbol ``\texttt{@}'' is not
recognized correctly in the Czech invoices with the output being
recognized as ``\texttt{\&}'', ``\texttt{\&\&}'', ``\texttt{Q}'', or ``©''.  Another
typical OCR error is a duplicated number. For example, in the VAT
number ``\texttt{CZ\underline{00}176150}'' the ``\texttt{0}'' is
doubled to ``\texttt{CZ\underline{000}176150}''. 
Besides OCR errors, a missing part of address is the reason for 
46\% of partial matches within the English invoices and for 16\% in the Czech invoices
respectively. The extraction module only looks for (parts of) addresses in continuous lines.
Therefore, in cases where some other pieces of information are inserted in those lines, e.g.\
a VAT number, a part of the address is misidentified.
An overview of all
the error types is presented in Table~\ref{tab:error_types}.

\section{Conclusions and future directions}

In this paper, we have presented the current results of the OCRMiner
system for extracting information from semi-structured scanned
business documents. OCRMiner is designed to process the document in
the same way as a human reader employing a combination of text
analysis techniques and positional layout features. 

The system has been evaluated with two invoice datasets (English and
Czech) containing data from more than 150 different vendors from all
over the world.  Using an open source OCR, with a small development
set consisting of only 20 invoices, the system is able to detect
\textbf{90.5\%} items in the English invoices and \textbf{88.2\%} in
the Czech invoices with 93--96\% of these item values being
{correct} matches. OCRMiner is also able to detect whether an image is
(the first page of) an invoice or not with \textbf{98\%} accuracy.

Such result is fully comparable to the
state-of-the-art systems which are, however, trained on much larger
datasets. The main advantages of OCRMiner lie in employing rich text
analysis annotations combined with the positional properties of each
text block. With underlying language dependent modules (OCR, keywords
and named entities), the higher-level modules are language independent
and offer a straightforward application to a new language without
a need of huge annotated invoice datasets. The system is also able to
cope with OCR errors using similarity search keyword detection.

The system tries to mimic the extraction procedure used by humans 
when reading an invoice. Even when the layout changes, 
the principle that the data is located around keywords 
(often on the right or under, or in some cases, above) 
should not change. This allows OCRMiner to adapt to most 
possible new layout varieties.
Although, as is mentioned in the error analysis, 
the rules have limited effect in distinguishing the buyer 
and seller blocks when there are no keywords attached 
to them. In this case, the rules cover the most 
frequent possibilities, but with some invoices they may 
even conflict with the designer's decision. For example, 
the seller mostly appears on the left of the buyer but, 
in some rare cases, the seller and buyer position are swapped.
In the future version, an ensemble seller/buyer distinguishing model
with a supervised learning method will be tested as preliminary
experiments show promising results.

Another possibility to improve the performance lies in 
the layout analysis step. A portion of errors are 
caused by block fragmentations where the desired data is 
disconnected from the keywords and the surrounding context. Identification
of such remote pieces of information as e.g.\ parts of the seller or
buyer address can further boost the accuracy of extracted information.
One possible direction is taking the advantage of
drawn tabular lines that separate invoice parts but that are ignored
by the OCR process.

In the ongoing research, OCRMiner is currently adapted to the
information extraction task from other types of business documents
such as contracts, reports, orders, or meeting minutes.

\begin{acknowledgements}
This work has been partly supported by the Ministry of Education of CR
within the LINDAT/\discretionary{}{}{}CLARIAH-CZ research
infrastructure LM2018101 and by Konica Minolta Business Solution Czech
within the OCRMiner project.
\end{acknowledgements}

\hbadness=10000
\bibliographystyle{splncs}
\bibliography{main}

\end{document}